%% file: sn-article.tex
\documentclass[referee,sn-vancouver-num]{sn-jnl}% Math and Physical Sciences Numbered Reference Style
\usepackage{tipa}
\usepackage{adjustbox}
\usepackage{graphicx}
\usepackage{setspace}
\usepackage[title]{appendix}
\usepackage{array}
\usepackage{float}
\usepackage{booktabs}
\usepackage[utf8]{inputenc} % Ensure correct character encoding
\usepackage{graphicx}%
\usepackage{multirow}%
\usepackage{amsmath,amssymb,amsfonts}%
\usepackage{amsthm}%
\usepackage{mathrsfs}%
\usepackage[title]{appendix}%
\usepackage{xcolor}%
\usepackage{textcomp}%
\usepackage{manyfoot}%
\usepackage{booktabs}%
\usepackage{algorithm}%
\usepackage{algorithmicx}%
\usepackage{algpseudocode}%
\usepackage{listings}%
\usepackage{subcaption}%
\usepackage{orcidlink}
\floatstyle{plaintop}
\restylefloat{table}
\theoremstyle{thmstyleone}%
\theoremstyle{thmstyletwo}%

\theoremstyle{thmstylethree}%

\begin{document}

\title[Dialect and Gender Bias in YouTube's Spanish Captioning System]{Dialect and Gender Bias in YouTube's Spanish Captioning System}


\author*[1]{\fnm{Iris Dania} \sur{Jimenez}\orcidlink{0009-0005-7803-0060}}
\email{iris.jimenez@campus.lmu.de}

\author[1,2]{\fnm{Christoph} \sur{Kern}\orcidlink{0000-0001-7363-4299}}%\email{christoph.kern@stat.uni-muenchen.de}

\affil[1]{\orgdiv{Department of Statistics}, 
\orgname{Ludwig-Maximilians-Universität München, Munich, Germany}}

\affil[2]{\orgname{Munich Center for Machine Learning (MCML), Munich, Germany}}

% YOUR CUSTOM TEXT BEFORE ABSTRACT

% 150 to 250 words
\abstract{Spanish is the official language of twenty-one countries and is spoken by over 441 million people. Naturally, there are many variations in how Spanish is spoken across these countries. Media platforms such as YouTube rely on automatic speech recognition systems to make their content accessible to different groups of users. However, YouTube offers only one option for automatically generating captions in Spanish. This raises the question: could this captioning system be biased against certain Spanish dialects? This study examines the potential biases in YouTube's automatic captioning system by analyzing its performance across various Spanish dialects. By comparing the quality of captions for female and male speakers from different regions, we identify systematic disparities which can be attributed to specific dialects. Our study provides further evidence that algorithmic technologies deployed on digital platforms need to be calibrated to the diverse needs and experiences of their user populations. Code avalaible \href{https://github.com/irisdaniaj/Spanish-Dialect-Bias-in-Youtube-s-Captioning-system}{here}.}




\maketitle

\newpage

\section{Introduction}

Spanish is the official language in 21 countries and is spoken by over 441 million people globally \cite{moreno2007atlas}. It holds a significant presence across various continents, including Europe, where it is the dominant language in Spain, and the Americas, where it is the primary language in most of Latin America. In the United States, Spanish is widely spoken as both a first and second language, particularly in states with large Hispanic populations such as California, Texas, and Florida. The language’s reach extends even to Africa, where it is the official language of Equatorial Guinea, and to Antarctica, where Chilean and Argentinian research stations maintain Spanish as a working language. Notably, Spanish is one of the six official languages of the United Nations \cite{spanish_un}. It is the second most spoken language by native speakers, the fourth most spoken language worldwide, and ranks as the third most used language on the internet \cite{spanish_intern_usage, eberhard2022summary}.

The Spanish language exhibits substantial regional variations, with major dialects such as Castilian, Mexican, Caribbean, Andean, Chilean, Paraguayan and Rioplatense \cite{hualde2005sounds}. These dialects differ in vocabulary, pronunciation, and grammar, posing challenges for standardization in media and technology. Recognizing and accommodating these differences is crucial for technology platforms, particularly those that rely on speech recognition and text generation, to ensure accuracy and accessibility for all Spanish speakers. The challenge lies in developing systems that can accurately interpret and transcribe speech from speakers of different dialects without losing the nuances that make each dialect unique. This is particularly important for educational tools, automated customer service systems, and any application where clear communication is essential.

YouTube, as a leading global platform for content creation and consumption, serves billions of users worldwide and has a unique role in facilitating language learning and cultural exchange. With over two billion logged-in monthly users \cite{youtube2023summary}, YouTube has become an essential resource for a wide array of activities. These range from educational tutorials, where users can learn new skills or languages, to entertainment, news and political content, where users stay informed or may engage in (polarized) discussions \cite{Grusauskaite2024, Jang2023, Chen2022}. The platform’s vast repository of content includes videos in countless languages, making it a versatile tool for learning and engagement for billion of users worldwide.
The platform's accessibility features, such as automatic captions, are vital for ensuring that its content is inclusive and available to all users, regardless of their language proficiency or hearing ability. Automatic captions help non-native speakers understand content in different languages, support individuals with hearing impairments, and enhance the overall accessibility of videos. These features are particularly important for educational and informational content, where accurate and accessible captions can significantly enhance comprehension and learning outcomes. 

Despite the extensive regional variations in the Spanish language, YouTube currently provides only a single option for generating Spanish captions \cite{youtube2024caps}. 
This approach may not account for the significant differences in vocabulary, pronunciation, and grammar across various Spanish dialects. As a result, the captions may not accurately reflect the spoken content for speakers of different dialects, leading to potential misunderstandings and reduced accessibility. 

This study evaluates the performance of YouTube’s captioning system across different Spanish dialects for male and female speakers. While its performance has been tested for different English dialects \cite{tatman-2017-gender}, we present the first study that analyzes YouTube's caption performance for both gender and Spanish dialects. We draw on extensive audio data which include seven different Latin American Spanish dialects, segmented by gender, and evaluate the accuracy of Youtube's generated captions using the Word Error Rate (WER), Character Error Rate (CER) and Entity Recall metrics. The results revealed notable disparities, with female Puerto Rican speakers achieving the lowest WER at 16\%, while Argentinian speakers experienced the highest WER at 24\%, indicating potential biases in the system’s ability to accurately transcribe certain dialects. By systematically analyzing how the system handles various dialects and genders, this research sheds light on whether the current captioning system adequately serves the diverse Spanish-speaking population or if there are gaps that need to be addressed. Understanding these biases is crucial for improving the accessibility and accuracy of automated captioning, ensuring that all users, regardless of their linguistic background, can fully benefit from media platforms such as YouTube.
Following unfolding research on bias in AI \cite{mitchell2021algorithmic, mehrabi2021, gerdon2022}, our study highlights that algorithmic technologies deployed on large-scale digital platforms need to be carefully assessed and curated as they provide critical connection points through which different user populations interact with digital content.

\section{Background}

\subsection{Spanish Dialects}

Spanish is a language rich in dialectal diversity, with significant variations across different regions. The major widely recognized dialects include Castilian, Mexican, Central American, Caribbean, Paraguayan, Chilean, Rioplatense and Andean Spanish, with each dialect exhibiting regional phonetic, lexical, and grammatical characteristics \cite{hualde2005sounds}.  These dialectal differences are not just linguistic pecularities but are deeply rooted in the historical colonization, migration patterns, and interactions with indigenous languages and other European influences over the centuries. 

Castilian Spanish, which is predominant in Spain, is often considered the standard form of the language in educational and media contexts. It is characterized by its use of the \textipa{/$\theta$/} sound for the letters 
\textless\textless c\textgreater\textgreater, \textless\textless z\textgreater\textgreater  before \textless\textless i\textgreater\textgreater, \textless\textless e\textgreater\textgreater
, a feature known as \textless\textless distinción\textgreater\textgreater. % This phonological feature sets Castilian apart from many other Spanish dialects and has become a symbol of Spanish identity within Spain.
In contrast, Latin American Spanish, which includes a broad array of regional variations, generally does not distinguish between the \textipa{/s/} and \textipa{/$\theta$/} sounds, a phenomenon known as \textless\textless seseo\textgreater\textgreater. This lack of distinction is one of the most prominent phonological differences between European and Latin American Spanish, highlighting how geographic separation and colonial history have led to divergent linguistic evolutions. \\
Mexican Spanish, mainly spoken in Mexico and the southern regions of the United States, is particularly interesting due to its incorporation of numerous indigenous terms, a reflection of the country’s rich pre-Columbian history \cite{hualde2005sounds}. Additionally, it features distinctive diminutives and the consonant cluster \textipa{/tl/}, inherited from the indigenous pre-Columbian language that profoundly influenced Mexican Spanish. This cluster is particularly challenging for speakers of other Spanish dialects to pronounce, underlining the unique phonetic inventory that has developed in Mexico over centuries \cite{hualde2005sounds}. The Central American Spanish is spoken in Guatemala, El Salvador, Honduras, Nicaragua and Costa Rica. In the central nations of El Salvador, Honduras, and Nicaragua, the \textipa{/s/} sound at the end of a syllable or before a consonant is often pronounced as \textipa{[h]}, though this is less common in formal speech such as TV broadcasts \cite{lipski2008central}. Caribbean Spanish is spoken in Cuba, Dominican Republic, Puerto Rico, Panama and the coasts of Venezuela and Colombia. It closely resembles the Spanish spoken in the Canary Islands and, to a lesser extent, the Spanish of western Andalusia. It is noted for its rapid speech and the aspiration or omission of the \textipa{/s/} sound at the end of syllables \cite{lipski2008central}. In Paraguay, Spanish coexists with Guarani, an indigenous language that has official status alongside Spanish. Paraguayan Spanish, also spoken in the lowlands of Eastern Bolivia \cite{hualde2005sounds}, exhibits distinctive features reminiscent of the Spanish formerly spoken in northern Spain. This is due to the significant number of early Spanish colonizers originating from the Basque Country that settled in the region. The Guarani language has greatly influenced Paraguayan Spanish, affecting both its vocabulary and grammar, leading to a unique linguistic blend that reflects the country’s dual linguistic heritage \cite{hualde2005sounds}. Chilean Spanish is primarily spoken in Chile and the neighbouring areas. The Royal Spanish Academy recognizes 2,214 words and idioms exclusively or mainly produced in Chilean Spanish, in addition to many still unrecognized slang expressions \cite{chilenismos2020}. Chilean Spanish is also notoriously known among Spanish native speakers to be one of the most different dialects \cite{alemany2021chileno}. Rioplatense Spanish, spoken in Argentina and Uruguay, is distinctive for its intonation, which often resembles the Neapolitan language of Southern Italy. This feature is a legacy of the massive Italian immigration to Argentina and Uruguay in the late 19th and early 20th centuries \cite{Zenkovich2018ParticularidadesDI}. The use of the pronouns \textless\textless vos\textgreater\textgreater instead of \textless\textless tú\textgreater\textgreater for informal address is another characteristic feature of Rioplatense Spanish, differentiating it from other Spanish dialects. Andean Spanish is a dialect spoken in the central Andes, stretching from southern Colombia to northern Chile and northwestern Argentina, and encompassing Ecuador, Peru, and Bolivia. This dialect, while similar to other forms of Spanish, is heavily influenced by indigenous languages such as Quechua and Aymara \cite{hualde2005sounds}.

These dialectal distinctions present challenges for standardizing the language and for technological applications like automatic speech recognition, which must accommodate this linguistic diversity to perform accurately and inclusively. Understanding and accounting for these regional variations is crucial for developing systems that can accurately transcribe and interpret spoken Spanish across different dialects.

\subsection{Automatic Speech Recognition Systems}

The development of Automatic Speech Recognition (ASR) systems has a rich history dating back to the 1950s. Initially, these systems were limited to recognizing isolated digits and small vocabularies \cite{ASRhistory}. One of the earliest notable examples was IBM's \textless\textless Shoebox\textgreater\textgreater from the 1960s, which could understand and respond to a small set of spoken commands. However, these early systems were far from perfect, often requiring speakers to enunciate clearly and pause between words to achieve any level of accuracy. These limitations highlighted the complexity of human speech and the challenges in developing systems that could mimic the natural language processing capabilities of the human brain.

Significant advancements in ASR technology occurred in the 1970s and 1980s with the introduction of Hidden Markov Models (HMMs). HMMs allowed ASR systems to model the temporal variations in speech, significantly improving accuracy by enabling the system to predict the probability of a sequence of phonemes rather than relying on static, isolated sounds \cite{jelinek1997statistical}. This advancement marked a pivotal shift from simple pattern recognition towards more sophisticated statistical modeling, paving the way for more complex and capable ASR systems.

By the 1990s, the field of ASR technology experienced further improvements with the integration of statistical language models and the rise of large-vocabulary continuous speech recognition (LVCSR) systems \cite{536824}. These systems were capable of understanding continuous speech, where words are spoken in a flow rather than in isolation, which is closer to how people naturally speak. This development was made possible by the increasing computational power available at the time, which allowed for the processing of larger datasets and more complex algorithms. Additionally, the availability of large-scale annotated corpora enabled the training of more robust models that could handle the variability inherent in human speech, such as differences in accent, intonation, and speaking style.

In recent years, ASR technology has seen remarkable advancements, particularly in languages like English and Chinese. These improvements have been driven by extensive research, the availability of vast datasets, and the application of sophisticated machine learning algorithms, particularly deep learning techniques. For example, Google's ASR system has achieved high accuracy rates in English, largely due to the availability of large, diverse datasets and continuous technological enhancements \cite{saon2016ibm2016englishconversational}. The system's success can be attributed to its ability to learn from a vast amount of data, encompassing a wide range of accents, speech patterns, and contextual uses of language. Similarly, Chinese ASR systems have benefited from targeted research efforts and the integration of tonal and phonetic elements unique to the language, leading to robust and reliable performance \cite{tonale}. The success of these systems in handling languages with complex tonal and phonetic structures showcases the versatility and adaptability of modern ASR technologies.

Despite these advancements, modern ASR systems are not error-free and can show significant performance gaps between different dialects of a language. Recent studies have demonstrated that the performance of ASR systems often declines when tested on dialectal variations of a language that were not included in the training data \cite{ELFEKY20181, chan22b_interspeech}. This decline in performance highlights the importance of training ASR systems on diverse datasets that include a wide range of dialects and speaking styles. However, the challenge remains that training such systems on a multitude of dialects can dilute their effectiveness for any single dialect, as models trained on multiple dialects at once tend to be less effective than those specifically trained for individual dialects \cite{parsons2023a, chan22b_interspeech}. The Spanish language is spoken by over 441 million people across 21 countries, each with its own regional dialects and variations in pronunciation, vocabulary, and syntax. This diversity poses a significant challenge for ASR systems, which must be able to accurately recognize and transcribe speech from speakers with different linguistic backgrounds.

The issue of dialect bias in ASR systems has been explored in selected languages beyond Spanish. For instance, different studies have been conducted on dialect bias for Arabic \cite{arabic, Sawalha2013TheEO} and English \cite{WHEATLEY199145, tatman-2017-gender, markl2022language} speakers, as summarized in Table \ref{tab:ASR_overview}. These prior studies have found that ASR systems often perform better on the dialects or accents that are more prominently represented in the training data, leading to disparities in performance that can disadvantage speakers of less common or non-standard dialects. However, the literature still lacks a comprehensive study of Spanish dialect bias in ASR systems. This gap is significant given the widespread use of Spanish globally and the increasing reliance on ASR technology in everyday applications such as virtual assistants, automated transcription services, and language learning tools.

Nevertheless, considerable work has been done to improve current ASR systems or create new ones for the Spanish language. Efforts have included implementing a single multidialectal model to accommodate the diverse Spanish dialects spoken across Europe and Latin America \cite{CABALLERO2009217}. Additionally, researchers have developed regionalized NLP models for Spanish language variations based on Twitter data, which offers a rich and diverse source of linguistic information \cite{tellez2023regionalized}. Furthermore, there has been work on creating automatic dialect recognizer systems specifically for Mexican \cite{hernandez2017automatic}, Cuban, and Peruvian Spanish \cite{543236}. These recognizer systems aim to improve the accuracy of ASR systems by tailoring them to specific dialects, thereby reducing the errors that arise from dialectal variation. Other efforts have focused on improving the resilience of ASR models against different native Spanish accents \cite{chitkara2022pushingperformancesasrmodels} and performing punctuation restoration for speech-to-text ASR systems \cite{zhu2022punctuationrestorationspanishcustomer}. These advancements reflect a growing recognition of the need to address linguistic diversity in ASR systems and the ongoing efforts to make these systems more inclusive and accurate for all users. While there is promising research aimed at improving ASR systems, there remains a critical need for comprehensive audits of large-scale, widely deployed systems, such as YouTube's ASR. 

So far, this discussion has primarily focused on the biases that ASR systems can exhibit based on different dialects. However, in the literature, there has been a significant amount of work focusing on the gender of the speakers as well. Male and female voices have different acoustic characteristics, such as pitch, tone, and speech patterns \cite{gender}. These differences have been shown to affect the performance of ASR systems, with studies demonstrating bias against female speakers \cite{garnerin2019genderrepresentationfrenchbroadcast} and, in some cases, bias against male speakers \cite{Sawalha2013TheEO, feng2021quantifyingbiasautomaticspeech, AddaDecker2005DoSR}. These studies reveal that while ASR systems are improving, they are not yet fully equitable across all demographics. 

Furthermore, ASR systems have also been shown to work better for younger speakers compared to older speakers \cite{Sawalha2013TheEO}. This age-related bias likely stems from the fact that most training data for ASR systems comes from younger adults, leading to a system that is better tuned to the speech patterns of this demographic. Additionally, studies have shown that ASR systems exhibit an uneven performance when comparing English white speakers and English African American speakers, with worse performances for the latter group \cite{koenecke2020racial}. This racial bias in ASR systems has serious implications for their use in real-world applications, particularly in areas such as law enforcement and customer service, where accurate speech recognition is crucial. Furthermore, speech disabilities can impact the performance of ASR systems, with research showing that individuals with speech impairments often experience higher error rates when using these technologies \citep{moro2019study, halpern2020detecting}. 
% These findings underscore the need for more inclusive training data that represents a wider range of speech patterns and the development of ASR systems that are robust to variations in speech.

ASR systems have thus shown disparities in performance across various attributes, including dialects, gender, age, and speech disabilities. Extensive work has been done to study these biases, but critical gaps remain. 
The most similar work to this study is the research by Tatman \cite{tatman-2017-gender}, which tested the quality of YouTube's automatically generated captions by gender and for different English dialects in 2017. Higher error rates were detected for women and speakers from Scotland, indicating that even widely used systems like YouTube’s ASR face critical challenges (see Table \ref{tab:ASR_overview}). Next to focusing on Spanish and YouTube's current ASR system, we go one step further and aim to explain performance variations based on prosodic features of the recorded voice data. Another relevant study is \cite{ELFEKY20181}, where the Google Assistant Voice system was tested for five different Spanish dialects (US, Spain, Mexico, Argentina, and Latin America). The study found that the system's performance varied across these dialects, further highlighting the need for more targeted improvements in ASR systems. To the best of our knowledge, however, there has not yet been a study that analyzes YouTube's captioning system performance across both Spanish dialects and gender. 

\onehalfspacing

\begin{table}[H]
\centering
\scriptsize
\caption{Summary of ASR's (by YouTube, Google, Amazon, and custom-build) performance on different dialects}
\vspace{0.3em}
\label{tab:ASR_overview}
\small
\begin{subtable}{0.5\textwidth}
\centering
\subcaption{Summary of Youtube's ASR performance\\ on English dialects from \cite{tatman-2017-gender}}
\begin{tabular}{l|ccc}
\hline
\textbf{Region} 
 & \textbf{WER(F)\,$\downarrow$} 
 & \textbf{WER(M)\,$\downarrow$} \\
\hline
California   & 0.38 & 0.30 \\
New Zealand   & 0.60 & 0.27 \\
New England   & 0.49 & 0.37\\
Georgia      & 0.56 & 0.37 \\
Scotland     & 0.52 & 0.55 \\
\hline
\textbf{Gender} \\ 
\hline
Female      & 0.51  \\
Male        & 0.37  \\ 
\hline
\end{tabular}
\end{subtable}%
\begin{subtable}{0.5\textwidth}
\centering
\subcaption{Performance of the ASR model created by the authors of \cite{arabic} on different Arabic dialects}
\begin{tabular}{l|ccc}
\hline
\textbf{Region} 
 & \textbf{Accuracy}  \\ 
\hline
Algiers     & 98.87 \%\\
Bechar      &  78.26 \%\\
El-Oued     & 76.60\% \\
Ghardaia    & 67.44\%   \\
Jijel       & 100 \%\\
Tizi-Ouzou  & 97.96\% \\
\hline
\end{tabular}
\end{subtable}

\begin{subtable}{1\textwidth}
\centering
\subcaption{Summary of Google Speech-to-Text performance on English dialects from \cite{markl2022language}}
\begin{tabular}{l|ccc}
\hline
\textbf{Dialect} 
 & \textbf{WER(F) $\downarrow$} 
 & \textbf{WER(M) $\downarrow$}  \\ 
\hline
Belfast            & 0.19 & 0.21 \\
Bradford Punjabi   & 0.19 & 0.22 \\
Cambridge          & 0.16 & 0.11 \\
Cardiff Welsh      & 0.18 & 0.14 \\
Dublin             & 0.12 & 0.17 \\
Leeds              & 0.9  & 0.16 \\
Liverpool          & 0.17 & 0.17 \\
London West Indian & 0.10 & 0.18 \\
Newcastle          & 0.17 & 0.15 \\
\hline
\end{tabular}
\end{subtable} 

\begin{subtable}{1\textwidth}
\centering
\subcaption{Summary of Amazon Transcribe performance on English dialects from \cite{markl2022language}}
\begin{tabular}{l|ccc}
\hline
\textbf{Dialect} 
 & \textbf{WER(F) $\downarrow$} 
 & \textbf{WER(M) $\downarrow$}  \\ 
\hline
Belfast            & 0.56 & 0.34 \\
Bradford Punjabi   & 0.40 & 0.33 \\
Cambridge          & 0.40 & 0.37 \\
Cardiff Welsh      & 0.32 & 0.28 \\
Dublin             & 0.30 & 0.44 \\
Leeds              & 0.31  & 0.31 \\
Liverpool          & 0.33 & 0.31 \\
London West Indian & 0.22 & 0.21 \\
Newcastle          & 0.39 & 0.26 \\
\hline
\end{tabular}
\end{subtable}
\end{table}

\doublespacing

\section{Data}

For this work, two datasets are used; the Crowdsourcing Latin American Spanish for Low-Resource Text-to-Speech dataset \cite{guevara-rukoz-etal-2020-crowdsourcing} for the Argentinian, Chilean, Colombian, Peruvian, Puerto Rican and Venezuelan dialects, and the TEDx Spanish Corpus \cite{mena_2019} for the Mexican dialect. 

The Crowdsourcing Latin American Spanish for Low-Resource Text-to-Speech dataset consists of crowdsourced recordings from both male and female speakers, along with their corresponding orthographic transcriptions. Any mismatches between transcriptions and audio were human-corrected during quality control.
All recorded volunteers were native speakers of their respective dialects. Recordings for the Argentinian, Chilean, Colombian, and Peruvian dialects were conducted in their native regions, while recordings for the Puerto Rican and Venezuelan dialects took place in New York, San Francisco, and London. 
The original recording script, designed for Mexican Spanish, was adapted for the different dialects by shortening phrases and removing references specific to Mexican Spanish. The script included approximately 30 ``canonical'' sentences across all dialects to capture phonological contrasts. Additional sentences were generated using templates to increase variety. % Dialect-specific pronunciation lists were expanded to cover more words, with manual edits to correctly capture the pronunciation of loanwords%. 
Although only a small portion of the script was specifically adapted by native speakers for each dialect, speakers were allowed to improvise to ensure a natural representation of their dialects. In any of such cases, the transcriptions were subsequently updated to match the spoken content. 
In total, the recordings for the Latin American dataset comprise 19 hours of female speakers and 18 hours of male speakers, totaling 37 hours of audio with 176 unique speakers. For Puerto Rico, only female speakers were recorded. The audios were recorded as 48 kHz single-channel and are provided in 16 bit linear PCM RIFF format. 

The TEDx Spanish Corpus is a 24-hour, gender-imbalanced dataset featuring spontaneous speeches from various Mexican TEDx event presenters. This dataset contains 11243 audio files from 142 different speakers, 102 of whom are male and 40 female. Audios and transcriptions are provided in lowercase without punctuation, and all the transcriptions were done by native Spanish speakers. The audio files are distributed in Windows WAVE 16 kHz @ 16-bit mono format.

For each dataset, every audio recording was assigned a unique identifier to distinguish between multiple recordings from the same speaker. This organization was necessary because each speaker contributed more than one audio sample. 
% In total, combining both the datasets, a collection of 69 hours of audio was gathered.

\section{Methods}

This study evaluates the performance of YouTube's automatic captioning system for male and female speakers across different Spanish dialects. % The methodology involved addressing technical challenges related to audio uploads, synchronization issues, and analyzing and modeling prosodic features and caption accuracy.
We use the outlined audio data to generate captions using Youtube's Data API v3 and compare the retrieved captions with the human-generated transcripts of both data sources using various measures (see below). 

To streamline the process, audio files were grouped by gender and dialect into longer concatenated files. Timestamp mapping files were created to preserve segment-level information, where each segment corresponds to a sentence. A five-second pause was added between audio segments to improve alignment with the ground truth transcriptions. Since the original audio files had varying sampling rates, all files were first converted to a standard 16 Hz sampling rate to ensure consistency before calculating prosodic features. The concatenated audio files were then split into thirty-minute blocks to meet YouTube's computational constraints. After uploading, captions for each video (i.e., thirty-minute audio blocks with a black screen) were retrieved using YouTube's API and compared against the ground truth transcriptions on a sentence-by-sentence basis. A summary of the data analyzed in this study is presented in Table \ref{tab1}.

Caption accuracy is assessed using three metrics: Word Error Rate (WER), Character Error Rate (CER) and Entity Recall as defined in Appendix~\ref{appendix}, which measure transcription quality at different levels. 

The study further analyzes two key prosodic features: average pitch and intensity. Pitch, measured as the fundamental frequency (F0), typically reflects gender differences, with male speakers exhibiting lower values due to anatomical differences \cite{zhang2021contribution}. It can also reveal dialectal variations, as pitch contours and ranges often differ across regions \cite{Keating2012ComparisonOS}. Intensity, measured as Root Mean Square (RMS) energy, reflects the loudness of speech. It can highlight cultural or regional norms, as some dialects are characterized by louder or softer speech patterns \cite{Chen2005}.

Based on the audio features and accuracy measures, we fit mixed-effects regression models (with sentences (level-1) nested in speakers (level-2)) to assess how different attributes contribute to errors in the retrieved captions. The analysis includes a total of 278 speakers and 2,147 sentences. The mixed-effects models focus on WER as the outcome variable and include random intercepts as individual speakers could contribute multiple audio recordings. The independent variables include pitch, intensity, country, and gender. We estimate multiple models to systematically evaluate factors influencing WER while controlling for speaker-specific effects.

\begin{table}[ht]
\centering
\caption{Summary of processed data with YouTube captions.}
\label{tab1}
\begin{tabular}{@{}lcrrr@{}}
\toprule
\textbf{Country} & \textbf{Gender} & \textbf{Duration (min)} & \textbf{Speakers} & \textbf{Sentences} \\
\midrule
Argentina  & F & 28:40 & 30 & 177\\
             & M & 28:42 & 13 & 183\\
Chile      & F & 28:48 & 13 & 164\\
             & M & 28:50 & 18 & 167\\
Colombia    & F & 28:48 & 16 & 167\\
             & M & 28:45 & 17 & 171\\
Mexico      & F & 29:56 & 36 & 139\\
             & M & 30:04 & 69 & 142\\
Peru     & F & 28:52 & 18 & 158\\
             & M & 28:45 & 20 & 161\\
Puerto Rico & F & 28:50 &  5 & 162\\
Venezuela   & F & 28:38 & 11 & 173\\
             & M & 28:39 & 12 & 183\\
\midrule
\textbf{Total:} & & 6h27min & 278 & 2147\\
\bottomrule
\end{tabular}
\end{table}

Code for replication purposes is available on \href{https://github.com/irisdaniaj/Spanish-Dialect-Bias-in-Youtube-s-Captioning-system}{GitHub}.

\section{Results}

In this section, the results of the analysis will be presented, beginning with an examination of the quality of the captions generated for each audio sample by using the case sensitive Word Error Rate (WER), Character Error Rate (CER) and Entity Recall. For case insensitive metrics please refer to Table \ref{tab:results_insensitive_wer} in the Appendix. Finally, we present the results of the mixed-effects models, which examines the influence of pitch, intensity, country, and gender on WER, while accounting for intra-speaker variability. Different combinations of these variables are assessed to study their relationships with caption accuracy.

From Table \ref{tab:updated_results_wer}, it is evident that YouTube's ASR system performs best for female Puerto Rican speakers, with a WER of 16\%, indicating that 84\% of captions accurately match the ground truth across all sentences for this group. Following this, Mexican Spanish exhibit an overall WER value of 18\%. In contrast, Argentinian Spanish yields the highest overall WER at 24\%, while Chilean, Colombian, Peruvian and Venezuelan Spanish show intermediate values of 20\% to 22\%.

The trends in CER generally align with WER but provide additional insights into finer transcription errors. Mexico, despite its relatively low WER (18\%), exhibits the highest CER of 10.5\%, suggesting that while the system captures words accurately, it may struggle with smaller transcription details at the character level. Venezuela, on the other hand, shows a high WER of 22\% and a correspondingly high CER of 9.2\%, reflecting significant challenges in accurately transcribing this dialect. For Argentina, which has the poorest WER performance at 24\%, the CER is 7.4\%, indicating stronger errors at the word level. Puerto Rico achieves the lowest CER of 5.8\% (for female speakers), further solidifying its position as the best-performing dialect.

The Entity Recall metric is consistent with the WER and CER results, showing higher (i.e., better) values for Mexico and Puerto Rico, and lower values for Argentina and Venezuela, with a notable distance between the best (Mexico: 0.91) and worst (Argentina: 0.81) performing dialect. Similar patterns can be observed for case insensitive Entity Recall (Table \ref{tab:results_insensitive_wer}). 

These results are noteworthy given the complexity of some dialects. For example, the Chilean dialect, often cited as one of the most challenging due to its rapid speech and colloquialisms \cite{alemany2021chileno}, does not result in the poorest performance. On the other hand, the good performance for (female) Puerto Rican speakers may be unexpected, as Caribbean Spanish typically features faster speech. This could be explained by Puerto Rico's association with the United States, potentially leading to greater representation in YouTube's training data.
The performance gap in overall WER between Mexico (18\%) and Argentina (24\%) highlights clear disparities across dialects. CER reveals that Peru's transcription accuracy at the character level is considerably better (7.2\%) compared to other regions such as Venezuela (9.2\%) or Mexico (10.5\%). These differences indicate that phonetic and lexical variability can affect both word-level and character-level performance, suggesting that the ASR model may not fully capture these nuances.

Analyzing WER and CER by gender reveals small differences on average: female speakers achieve a WER of 20\% and CER of 9.5\%, while male speakers have a WER of 21\% and CER of 8.5\%. Both female and male speakers also achieve comparable Entity Recall values overall. However, when stratified by both country and gender, the results show regional variations. For instance, in Chile and Mexico, male speakers exhibit higher WERs than females, and in Peru higher CER values, suggesting greater transcription challenges for male voices. Conversely, in Venezuela and Argentina, male speakers achieve lower WERs and CERs, and in Colombia lower CERs, indicating better performance in these regions. While these differences in WER are still relatively small, differences in CER are more pronounced. No data was available for male Puerto Rican speakers, limiting conclusions for this dialect.

\begin{table}[h]
\centering
\caption{Case sensitive WER, CER, and Entity Recall by country}
\label{tab:updated_results_wer}
\small
\begin{subtable}{1\textwidth}
\centering
\subcaption{Overall}
\begin{tabular}{l|ccc}
\hline
\textbf{Country} 
 & \textbf{WER\,$\downarrow$} 
 & \textbf{CER\,$\downarrow$} 
 & \textbf{Entity Recall\,$\uparrow$} \\ 
\hline
Argentina   & 0.24 & 0.074 &  0.81 \\
Chile       & 0.22 & 0.079 &   0.84 \\
Colombia    & 0.22 & 0.073 &    0.84 \\
Mexico       & 0.18 & 0.105 &   0.91 \\
Peru        & 0.20 & 0.072 &    0.85 \\
Puerto Rico* & 0.16 & 0.058 & 0.89 \\
Venezuela   & 0.22 & 0.092 &    0.82 \\ 
\hline
\textbf{Gender} & & & \\ 
\hline
Female      & 0.20 & 0.095 &  0.87  \\
Male        & 0.21 & 0.085 &  0.86  \\ 
\hline
\end{tabular}
\end{subtable}

\begin{subtable}{1\textwidth}
\centering
\subcaption{Female Speaker}
\begin{tabular}{l|ccc}
\hline
\textbf{Country} 
 & \textbf{WER(F)} 
 & \textbf{CER(F)} 
 & \textbf{Entity Recall(F)} \\ 
\hline
Argentina   & 0.24 & 0.087 & 0.81 \\
Chile       & 0.21 & 0.082 & 0.84 \\
Colombia    & 0.22 & 0.079 & 0.84 \\
Mexico      & 0.19 & 0.108 & 0.91 \\
Peru        & 0.20 & 0.065 & 0.85 \\
Puerto Rico & 0.16 & 0.058 & 0.89 \\
Venezuela   & 0.23 & 0.122 & 0.83 \\ 
\hline
\end{tabular}
\end{subtable}

\begin{subtable}{1\textwidth}
\centering
\subcaption{Male Speaker}
\begin{tabular}{l|ccc}
\hline
\textbf{Country} 
 & \textbf{WER(M)} 
 & \textbf{CER(M)} 
 & \textbf{Entity Recall(M)} \\ 
\hline
Argentina   & 0.23 & 0.061 & 0.81 \\
Chile       & 0.22 & 0.077 & 0.83\\
Colombia    & 0.22 & 0.066 & 0.84\\
Mexico      & 0.20 & 0.102 & 0.89 \\
Peru        & 0.20 & 0.080 & 0.84\\
Puerto Rico & -    & -     &  -\\
Venezuela   & 0.21 & 0.060 &  0.82\\ 
\hline
\end{tabular}
\end{subtable}
\end{table}

We next present the results of regression model comparisons, focusing on the key predictors of Word Error Rate, including intensity, pitch, country, and gender. The analysis highlights the best-performing models, the significance at 5\% of individual predictors, and their combined effects on ASR performance. To account for the effect of gender on pitch, whenever the two variables were present in a model, an interaction term between the two was added. Descriptive statistics for the prosodic features (intensity and pitch) by dialect are presented in Table \ref{tab:prosodic} in the Appendix. For an extended summary of all models, refer to Table \ref{tab:all_models} in the Appendix.

The model comparisons in Table \ref{tab:model_results} reveal that Model 7, with the lowest AIC of -1782.64, is the best-performing model, indicating that speech intensity and gender are the most significant predictor of WER. While models incorporating multiple predictors (e.g., Model 10: country, pitch and intensity) and interactions (e.g., Model 6: gender, pitch and its interaction) improve upon single-variable models, their performance is marginally inferior to the simpler intensity and gender only model. Country effects are significant for specific dialects, i.e. Puerto Rico and Peru, where the negative effects on WER for those countries (compared to Mexican Spanish as the reference category) suggest better ASR adaptation to these dialects. Gender alone has no significant impact on WER, suggesting that Youtube ASR system does not show a significant different performance for female and male Spanish speakers overall. The comparison between the best model (Model 7) and the full model (Model 12) shows a modest AIC difference, suggesting that including additional predictors and interaction terms does not substantially improve model performance. Overall, the results emphasize the importance of acoustic features like intensity, but also gender and specific country effects in predicting the WER of the retrieved YouTube captions.

\begin{table}[h]
\centering
\caption{Summary of models predicting Word Error Rate.}
\label{tab:model_results}
\begin{tabular}{lccc}
\hline
\textbf{Model} & \textbf{Predictors}           & \textbf{AIC}     & \textbf{Significant Variables} \\ \hline
1              & Country                      & -1537.32        & Puerto Rico, Peru              \\
2              & Gender                       & -1509.96        & None                           \\    
3              & Country + Gender             & -1535.51        & Puerto Rico                    \\
4              & Country + Pitch              & -1515.24        & Pitch, Puerto Rico             \\
5              & Country + Intensity          & -1780.80        & Intensity                      \\
6              & Gender + Pitch + Pitch:Gender & -1520.25       & Pitch                          \\
7              & Gender + Intensity           & \textbf{-1782.64}        & Intensity                      \\
8             & Country + Gender + Pitch + Pitch:Gender & -1547.42 & Pitch, Puerto Rico \\
9             & Country + Gender + Intensity & -1779.22        & Intensity                      \\
10             & Country + Pitch + Intensity  & -1779.54        & Intensity                      \\
11             & Gender + Pitch + Intensity + Pitch:Gender & -1779.07 & Intensity        \\
12             & Full Model (All Predictors + Pitch:Gender)  & -1776.02        & Intensity                      \\ \hline
\end{tabular}
\end{table}

\section{Discussion}

This study offers insights into the performance of YouTube’s ASR system across various Spanish dialects. The findings indicate that while the system performs relatively well for Puerto Rican female speakers, it also reveals notable disparities, such as a higher Word Error Rate (WER) for Argentinian speakers. These WER variations appear to be primarily influenced by speaker intensity next to regional dialect characteristics.

Comparing our results with those of Tatman et al. \cite{tatman-2017-gender}, the error variation of YouTube's captions between Spanish dialects is less pronounced than the variation observed between English dialects. In our study, the largest difference in WER between two Spanish dialects (Puerto Rico and Argentina) is 8\%. In contrast, \cite{tatman-2017-gender} reported a WER difference of 25\% between Scottish and Californian English dialects. This suggests that YouTube's ASR system exhibits less variability across Spanish dialects. However, YouTube may have updated and improved its ASR system since Tatman et al. conducted their study.

In summary, it is important to note that our analysis focused on only seven Spanish-speaking countries, while there are many more countries where Spanish is the official language, each with its own unique dialectal variations. Expanding the scope of this study to include a wider range of dialects would offer a more comprehensive understanding of YouTube's ASR system performance and help identify areas that require improvement.
Moreover, a more detailed analysis of the specific phonemes and linguistic features that challenge YouTube's captioning system would be highly valuable, similarly to what has been done in previous studies  for Dutch and Chinese Mandarin \cite{e050b066097e4e7aaa1ea8fb04cfb73a}. This phoneme-level analysis could identify specific sounds or combinations of sounds that are prone to errors, which could then be targeted for improvement in future versions of the ASR system. 
In addition to exploring gender and dialectal biases, it would be highly valuable to investigate how YouTube’s ASR system performs across different age groups. Age-related biases are a known issue in speech recognition technology, as the acoustic characteristics of speech can vary significantly across different stages of life. 
Furthermore, there is little research on the system's performance for non-binary or transgender speakers, highlighting a research gap that needs to be addressed to ensure ASR systems are inclusive for all users.
One technical limitation of this work were the difficulties to fully exploit the available datasets due to computational constraints and limitations imposed by YouTube's API. These constraints prevented the use of additional audio data beyond the 30-minute segments in our processed sample and might have enabled additional analyses.

A promising avenue for future research involves conducting longitudinal studies to monitor the performance of YouTube’s ASR system over time. We further advocate for cross-platform comparisons with other widely used platforms, such as Google Assistant Voice, Apple's Siri, or Amazon's Alexa.

\section{Conclusion}

We systematically evaluated the performance of YouTube's ASR system across different Spanish dialects, with an additional focus on gender-based differences. While the system demonstrated relatively strong performance for Puerto Rican female speakers, the results also highlighted considerable disparities, such as a higher WER for Argentinian speakers. Variation in WER values can be attributed to both prosodic features and regional dialects. These findings underscore the need for further research to better understand the linguistic and technical factors that influence Youtube's ASR performance. Our study adds to a body of literature on (subgroup) bias in AI \cite{mitchell2021algorithmic, mehrabi2021}, highlighting that algorithmic technologies deployed on digital platforms need to be carefully tailored to the diverse user populations interacting with the platform's content. Future studies should aim to broaden the scope of analysis to include a wider range of dialects and demographics, such as different age groups and underrepresented regions. Additionally, exploring phoneme-level challenges and conducting cross-platform comparisons will be crucial in identifying specific areas for improvement. 

\clearpage

\section*{Acknowledgements}
We would like to express our gratitude to Ailin Liu for her support in analyzing the prosodic features.

\section*{Conflict of Interest}
On behalf of all authors, the corresponding author states that there is no conflict of interest.

\section*{Data Availability Statement}
Mexican Spanish dataset from \cite{mena_2019} available \href{https://www.openslr.org/67/}{here}. \\
\href{https://www.openslr.org/61/}{Argentinian}, \href{https://www.openslr.org/71/}{Chilean}, \href{https://www.openslr.org/72/}{Colombian}, \href{https://www.openslr.org/73/}{Peruvian}, \href{https://www.openslr.org/74/}{Puerto Rican}, \href{https://www.openslr.org/75/}{Venezuelan} dataset from \cite{guevara-rukoz-etal-2020-crowdsourcing}.

\bibliography{sn-bibliography.bib}
\clearpage

\appendix
\input{appendix}

\end{document}

%% file: appendix.tex
\section{Appendix}
\label{appendix}

\subsection{Metrics}

We calculated Word Error Rate (WER), Character Error Rate (CER) and Entity Recall as follows: 

\begin{equation}
\text{WER} = \frac{\text{Substitutions} + \text{Deletions} + \text{Insertions}}{\text{Total Words in Ground Truth}}
\end{equation}

\begin{equation}
\text{CER} = \frac{\text{Substitutions} + \text{Deletions} + \text{Insertions}}{\text{Total Characters in Ground Truth}}
\end{equation}

\begin{equation}
\text{Entity Recall} = \frac{\text{Number of Words in Intersection of GE and CE}}{\text{Total Number of Words in GE}} 
\end{equation} 

With GE being the set of words present in the ground truth data and CE the set of words present in the retrieved captions.

The key distinction between the case sensitive and case insensitive versions of these metrics is whether the formulas consider case differences (e.g., "Word" $\neq$ "word"). 
% when evaluating substitutions, insertions, and deletions.

\subsection{Additional Results}

The results in Table \ref{tab:results_insensitive_wer} show variation in case-insensitive WER and Entiity Entity Recallby region: for example, Peru exhibits a relatively low WER (12,6\%) paired with a moderately high Entiity Entity Recall(90,6\%), while Mexico’s WER is higher (15,5\%) but its Entiity Entity Recall(93,5\%) is among the highest. Puerto Rico stands out for its extremely high Entiity Entity Recall(94,5\%). Gender differences are relatively small—female and male WER values (14,5\% vs. 14,6\%) and Entity Recalls (91,1\% vs. 91,2\%) are almost on par, indicating consistent performance regardless of speaker gender.

\begin{table}[h!]
\centering
\caption{Case insensitive WER, CER, and Entity Recall by country}
\label{tab:results_insensitive_wer}
\small
\begin{subtable}{1\textwidth}
\centering
\subcaption{Overall}
\begin{tabular}{l|ccc}
\hline
\textbf{Country} 
 & \textbf{WER\,$\downarrow$} 
 & \textbf{CER\,$\downarrow$} 
 & \textbf{Entity Recall\,$\uparrow$} \\ 
\hline
Argentina   & 0.154 & 0.055 & 0.87 \\
Chile       & 0.138 & 0.062 & 0.89 \\
Colombia    & 0.144 & 0.056 & 0.88 \\
Mexico      & 0.155 & 0.082 & 0.93 \\
Peru        & 0.126 & 0.055 & 0.90  \\
Puerto Rico & -     & -     & -     \\
Venezuela   & 0.145 & 0.075 & 0.88  \\  
\hline
\textbf{Gender} & & & \\ 
\hline
Female      & 0.145 & 0.075 & 0.91 \\
Male        & 0.146 & 0.065 & 0.90 \\ 
\hline
\end{tabular}
\end{subtable}

\begin{subtable}{1\textwidth}
\centering
\subcaption{Female Speaker}
\begin{tabular}{l|ccc}
\hline
\textbf{Country} 
 & \textbf{WER(F)} 
 & \textbf{CER(F)} 
 & \textbf{Entity Recall(F)} \\ 
\hline
Argentina   & 0.163 & 0.068 & 0.86 \\
Chile       & 0.132 & 0.064 & 0.90 \\
Colombia    & 0.145 & 0.064 & 0.88 \\
Mexico      & 0.156 & 0.086 & 0.92 \\
Peru        & 0.120 & 0.047 & 0.91 \\
Puerto Rico & 0.120 & 0.041 & 0.94 \\
Venezuela   & 0.156 & 0.105 & 0.88 \\ 
\hline
\end{tabular}
\end{subtable}

\begin{subtable}{1\textwidth}
\centering
\subcaption{Male Speaker}
\begin{tabular}{l|ccc}
\hline
\textbf{Country} 
 & \textbf{WER(M)} 
 & \textbf{CER(M)} 
 & \textbf{Entity Recall(M)} \\ 
\hline
Argentina   & 0.145 & 0.042 & 0.87 \\
Chile       & 0.144 & 0.059 & 0.89 \\
Colombia    & 0.143 & 0.048 & 0.88 \\
Mexico      & 0.153 & 0.079 & 0.92 \\
Peru        & 0.134 & 0.063 & 0.90 \\
Puerto Rico & -     & -     & -     \\
Venezuela   & 0.135 & 0.045 & 0.88 \\ 
\hline
\end{tabular}
\end{subtable}
\end{table}

From Table \ref{tab:prosodic}, it is evident that the Mexican Spanish audio files exhibit the highest values for both pitch and intensity, setting them apart from the other countries included in this study which show modest differences in pitch and intensity. Although all audio files were normalized, note that the data for Mexico was derived from a separate, distinct data source (i.e., TEDx talks).

When stratified by gender, the results align with findings from existing literature; as expected, female speakers exhibit higher average pitch values compared to male speakers (\cite{zhang2021contribution}). Intensity levels, on the other hand, are relatively consistent across genders, although slight variations are observed. For instance, Colombia shows higher intensity values for male speakers compared to females, while in other countries, intensity remains rather stable.

\begin{table}[h!]
\centering
\caption{Average pitch and intensity by country and gender.}
\label{tab:prosodic}
\small
\begin{tabular}{l|cc|cc|cc}
\hline
\textbf{Country}      & \textbf{Pitch} & \textbf{Intensity} & \textbf{Pitch (F)} & \textbf{Pitch (M)} & \textbf{Intensity (F)} & \textbf{Intensity (M)} \\ \hline
Argentina              & 144           & 0.022              & 171                & 117                & 0.021                  & 0.022                  \\
Chile                  & 158            & 0.021              & 197                & 118                & 0.021                  & 0.020                  \\
Colombia               & 146            & 0.024              & 178                & 114                & 0.021                  & 0.026                  \\
Mexico                 & 203            & 0.042              & 248                & 158                  & 0.043                  & 0.042                      \\
Peru                   & 147           & 0.024              & 171                  & 123                  & 0.023                      & 0.024                      \\
Puerto Rico            & -            & -              & 164                  & -                  & 0.028                      & -                      \\
Venezuela              & 159            & 0.026              & 196                  & 122                  & 0.026                      & 0.027                      \\ \hline
{\textbf{Gender}} \\ \hline
Female                 & 193            & 0.026              & -                  & -        & -                  & -           \\
Male                   & 126            & 0.027              & -                  & -        & -                  & -           \\ \hline
\end{tabular}
\end{table}

\begin{table}
\centering
\caption{Mixed-effects regression models predicting WER (ref.: Mexico (country) and Female (gender)). }
\label{tab:all_models}
\begin{subtable}[c]{1\textwidth}
\centering
\subcaption{\textbf{Model 1:} \(\texttt{word\_error\_rate} \sim \texttt{country}\)}
\begin{tabular}{lrrrrr}
\toprule
\textbf{Term} & \textbf{Coef.} & \textbf{Std.Err.} & \textbf{z} & \textbf{P>\textbar z\textbar} & \textbf{[0.025, 0.975]} \\
\midrule
Intercept                & 0.302 & 0.010 & 30.707 & 0.000 & [\,0.283, 0.321\,] \\
Chile       & -0.013 & 0.014 & -0.899 & 0.369 & [\,-0.041, 0.015\,] \\
Colombia     & -0.024 & 0.014 & -1.683 & 0.092 & [\,-0.052, 0.004\,] \\
Peru      & -0.029 & 0.014 & -2.076 & 0.038 & [\,-0.057, -0.002\,] \\
Puerto Rico & -0.080 & 0.021 & -3.756 & 0.000 & [\,-0.122, -0.038\,] \\
Venezuela    & -0.013 & 0.014 & -0.896 & 0.370 & [\,-0.041, 0.015\,] \\
Group Var                & 0.001 & 0.003 &        &       &                     \\
\midrule
\multicolumn{6}{l}{\textbf{AIC:} -1537.3232 \quad \textbf{BIC:} -1493.0751 \quad 
                   \textbf{Log-Likelihood:} 776.6616}\\
\multicolumn{6}{l}{$n_{Sentences}$: 2147 \qquad $n_{Speaker}$: 278} \\
\bottomrule
\end{tabular}
\end{subtable} \\
\begin{subtable}[c]{1\textwidth}
\centering
\subcaption{\textbf{Model 2:} \(\texttt{word\_error\_rate} \sim \texttt{gender}\)}
\begin{tabular}{lrrrrr}
\toprule
\textbf{Term} & \textbf{Coef.} & \textbf{Std.Err.} & \textbf{z} & \textbf{P>\textbar z\textbar} & \textbf{[0.025, 0.975]} \\
\midrule
Intercept      & 0.285 & 0.009 & 32.705 & 0.000 & [\,0.268, 0.302\,] \\
Male           & -0.000 & 0.013 & -0.032 & 0.974 & [\,-0.025, 0.024\,] \\
Group Var      & 0.004 & 0.014 &        &       &                     \\
\midrule
\multicolumn{6}{l}{\textbf{AIC:} -1509.9605 \quad \textbf{BIC:} -1487.8364 \quad 
                   \textbf{Log-Likelihood:} 758.9802}\\
\multicolumn{6}{l}{$n_{Sentences}$: 2147 \qquad $n_{Speaker}$: 278} \\
\bottomrule
\end{tabular}
\end{subtable} \\
\begin{subtable}[c]{1\textwidth}
\centering
\subcaption{\textbf{Model 3:} \(\texttt{word\_error\_rate} \sim \texttt{country + gender}\)}
\begin{tabular}{lrrrrr}
\toprule
\textbf{Term} & \textbf{Coef.} & \textbf{Std.Err.} & \textbf{z} & \textbf{P>\textbar z\textbar} & \textbf{[0.025, 0.975]} \\
\midrule
Intercept                & 0.304 & 0.011 & 28.501 & 0.000 & [\,0.283, 0.324\,] \\
Chile       & -0.013 & 0.014 & -0.878 & 0.380 & [\,-0.041, 0.016\,] \\
Colombia     & -0.024 & 0.014 & -1.668 & 0.095 & [\,-0.051, 0.004\,] \\
Peru      & -0.029 & 0.014 & -2.061 & 0.039 & [\,-0.057, -0.001\,] \\
Puerto Rico & -0.082 & 0.022 & -3.778 & 0.000 & [\,-0.125, -0.040\,] \\
Venzuela    & -0.013 & 0.014 & -0.878 & 0.380 & [\,-0.041, 0.016\,] \\
Male           & -0.004 & 0.009 & -0.429 & 0.668 & [\,-0.022, 0.014\,] \\
Group Var                & 0.001 & 0.003 &        &       &                     \\
\midrule
\multicolumn{6}{l}{\textbf{AIC:} -1535.5071 \quad \textbf{BIC:} -1485.7280 \quad 
                   \textbf{Log-Likelihood:} 776.7536}\\
\multicolumn{6}{l}{$n_{Sentences}$: 2147 \qquad $n_{Speaker}$: 278} \\
\bottomrule
\end{tabular}
\end{subtable} 
\end{table}

\begin{table}
\ContinuedFloat 
\centering
\begin{subtable}[c]{1\textwidth}
\centering
\subcaption{\textbf{Model 4:} \(\texttt{word\_error\_rate} \sim \texttt{country + pitch}\)}
\begin{tabular}{lrrrrr}
\toprule
\textbf{Term} & \textbf{Coef.} & \textbf{Std.Err.} & \textbf{z} & \textbf{P>\textbar z\textbar} & \textbf{[0.025, 0.975]} \\
\midrule
Intercept                & 0.358 & 0.028 & 12.912 & 0.000 & [\,0.304, 0.413\,] \\
Chile       & -0.014 & 0.020 & -0.706 & 0.480 & [\,-0.053, 0.025\,] \\
Colombia     & -0.027 & 0.019 & -1.388 & 0.165 & [\,-0.065, 0.011\,] \\
Peru      & -0.034 & 0.019 & -1.778 & 0.075 & [\,-0.072, 0.004\,] \\
Puerto Rico & -0.073 & 0.034 & -2.127 & 0.033 & [\,-0.140, -0.006\,] \\
Venezuela    & -0.014 & 0.021 & -0.663 & 0.507 & [\,-0.054, 0.027\,] \\
Pitch                    & -0.000 & 0.000 & -2.347 & 0.019 & [\,-0.001, -0.000\,] \\
Group Var                & 0.004 & 0.021 &        &       &                     \\
\midrule
\multicolumn{6}{l}{\textbf{AIC:} -1515.2448 \quad \textbf{BIC:} -1465.4656 \quad 
                   \textbf{Log-Likelihood:} 766.6224}\\
\multicolumn{6}{l}{$n_{Sentences}$: 2147 \qquad $n_{Speaker}$: 278} \\
\bottomrule
\end{tabular}
\end{subtable} \\
\begin{subtable}[c]{1\textwidth}
\centering
\subcaption{\textbf{Model 5:} \(\texttt{word\_error\_rate} \sim \texttt{country + intensity}\)}
\begin{tabular}{lrrrrr}
\toprule
\textbf{Term} & \textbf{Coef.} & \textbf{Std.Err.} & \textbf{z} & \textbf{P>\textbar z\textbar} & \textbf{[0.025, 0.975]} \\
\midrule
Intercept                & 0.461 & 0.015 & 30.309 & 0.000 & [\,0.431, 0.491\,] \\
Chile       & -0.022 & 0.017 & -1.310 & 0.190 & [\,-0.055, 0.011\,] \\
Colombia     & -0.010 & 0.017 & -0.603 & 0.547 & [\,-0.042, 0.022\,] \\
Peru      & -0.018 & 0.016 & -1.111 & 0.267 & [\,-0.050, 0.014\,] \\
Puerto Rico & -0.039 & 0.028 & -1.382 & 0.167 & [\,-0.095, 0.016\,] \\
Venezuela    & 0.018 & 0.018 &  1.031 & 0.303 & [\,-0.016, 0.052\,] \\
Intensity                & -7.253 & 0.456 & -15.901 & 0.000 & [\,-8.147, -6.359\,] \\
Group Var                & 0.003 & 0.005 &        &       &                     \\
\midrule
\multicolumn{6}{l}{\textbf{AIC:} -1780.7984 \quad \textbf{BIC:} -1731.0192 \quad 
                   \textbf{Log-Likelihood:} 899.3992}\\
\multicolumn{6}{l}{$n_{Sentences}$: 2147 \qquad $n_{Speaker}$: 278} \\
\bottomrule
\end{tabular}
\end{subtable} \\
\begin{subtable}[c]{1\textwidth}
\centering
\subcaption{\textbf{Model 6:} \(\texttt{word\_error\_rate} \sim \texttt{gender + pitch + pitch:gender}\)}
\begin{tabular}{lrrrrr}
\toprule
\textbf{Term} & \textbf{Coef.} & \textbf{Std.Err.} & \textbf{z} & \textbf{P>\textbar z\textbar} & \textbf{[0.025, 0.975]} \\
\midrule
Intercept             & 0.391 & 0.035 & 11.209 & 0.000 & [\,0.322, 0.459\,] \\
Male        & -0.020 & 0.063 & -0.321 & 0.748 & [\,-0.144, 0.104\,] \\
Pitch                 & -0.001 & 0.000 & -3.153 & 0.002 & [\,-0.001, -0.000\,] \\
Pitch:Male  & -0.000 & 0.000 & -0.298 & 0.766 & [\,-0.001, 0.001\,] \\
Group Var             & 0.004 & 0.012 &        &       &                     \\
\midrule
\multicolumn{6}{l}{\textbf{AIC:} -1520.2550 \quad \textbf{BIC:} -1487.0689 \quad
                   \textbf{Log-Likelihood:} 766.1275}\\
\multicolumn{6}{l}{$n_{Sentences}$: 2147 \qquad $n_{Speaker}$: 278} \\
\bottomrule
\end{tabular}
\end{subtable} \\
\begin{subtable}[c]{1\textwidth}
\centering
\subcaption{\textbf{Model 7:} \(\texttt{word\_error\_rate} \sim \texttt{gender + intensity}\)}
\begin{tabular}{lrrrrr}
\toprule
\textbf{Term} & \textbf{Coef.} & \textbf{Std.Err.} & \textbf{z} & \textbf{P>\textbar z\textbar} & \textbf{[0.025, 0.975]} \\
\midrule
Intercept       & 0.446 & 0.012 & 35.751 & 0.000 & [\,0.422, 0.470\,] \\
Male  & 0.009 & 0.010 &  0.890 & 0.374 & [\,-0.011, 0.030\,] \\
Intensity       & -7.154 & 0.447 & -15.990 & 0.000 & [\,-8.031, -6.277\,] \\
Group Var       & 0.002 & 0.004 &        &       &                     \\
\midrule
\multicolumn{6}{l}{\textbf{AIC:} -1782.6361 \quad \textbf{BIC:} -1754.9810 \quad
                   \textbf{Log-Likelihood:} 896.3181}\\
\multicolumn{6}{l}{$n_{Sentences}$: 2147 \qquad $n_{Speaker}$: 278} \\
\bottomrule
\end{tabular}
\end{subtable} 
\end{table}

\begin{table}
\ContinuedFloat 
\centering
\begin{subtable}[c]{1\textwidth}
\centering
\subcaption{\textbf{Model 8:} \(\texttt{word\_error\_rate} \sim \texttt{country + gender + pitch + pitch:gender}\)}
\begin{tabular}{lrrrrr}
\toprule
\textbf{Term} & \textbf{Coef.} & \textbf{Std.Err.} & \textbf{z} & \textbf{P>\textbar z\textbar} & \textbf{[0.025, 0.975]} \\
\midrule
Intercept                & 0.395 & 0.030 & 12.980 & 0.000 & [\,0.335, 0.455\,] \\
Chile       & -0.004 & 0.014 & -0.284 & 0.777 & [\,-0.032, 0.024\,] \\
Colombia     & -0.022 & 0.014 & -1.561 & 0.119 & [\,-0.050, 0.006\,] \\
Peru      & -0.027 & 0.014 & -1.949 & 0.051 & [\,-0.055, 0.000\,] \\
Puerto Rico & -0.087 & 0.021 & -4.041 & 0.000 & [\,-0.129, -0.045\,] \\
Venezuela    & -0.004 & 0.014 & -0.297 & 0.767 & [\,-0.033, 0.024\,] \\
Male           & 0.004 & 0.054 &  0.074 & 0.941 & [\,-0.102, 0.110\,] \\
Pitch                    & -0.001 & 0.000 & -3.220 & 0.001 & [\,-0.001, -0.000\,] \\
Pitch:Male     & -0.000 & 0.000 & -0.870 & 0.384 & [\,-0.001, 0.000\,] \\
Group Var                & 0.001 & 0.002 &        &       &                     \\
\midrule
\multicolumn{6}{l}{\textbf{AIC:} -1547.4167 \quad \textbf{BIC:} -1486.5755 \quad
                   \textbf{Log-Likelihood:} 784.7083}\\
\multicolumn{6}{l}{$n_{Sentences}$: 2147 \qquad $n_{Speaker}$: 278} \\                   
\bottomrule
\end{tabular}
\end{subtable} \\
\begin{subtable}[c]{1\textwidth}
\centering
\subcaption{\textbf{Model 9:} \(\texttt{word\_error\_rate} \sim \texttt{country + gender + intensity}\)}
\begin{tabular}{lrrrrr}
\toprule
\textbf{Term} & \textbf{Coef.} & \textbf{Std.Err.} & \textbf{z} & \textbf{P>\textbar z\textbar} & \textbf{[0.025, 0.975]} \\
\midrule
Intercept                & 0.458 & 0.016 & 29.230 & 0.000 & [\,0.428, 0.489\,] \\
Chile       & -0.023 & 0.017 & -1.374 & 0.169 & [\,-0.056, 0.010\,] \\
Colombia     & -0.011 & 0.017 & -0.652 & 0.514 & [\,-0.043, 0.022\,] \\
Peru      & -0.019 & 0.016 & -1.162 & 0.245 & [\,-0.051, 0.013\,] \\
Puerto Rico & -0.036 & 0.029 & -1.243 & 0.214 & [\,-0.092, 0.021\,] \\
Venezuela    & 0.017 & 0.018 &  0.975 & 0.329 & [\,-0.017, 0.052\,] \\
Male           & 0.008 & 0.011 &  0.747 & 0.455 & [\,-0.013, 0.030\,] \\
Intensity                & -7.278 & 0.456 & -15.943 & 0.000 & [\,-8.172, -6.383\,] \\
Group Var                & 0.003 & 0.005 &        &       &                     \\
\midrule
\multicolumn{6}{l}{\textbf{AIC:} -1779.2173 \quad \textbf{BIC:} -1723.9071 \quad
                   \textbf{Log-Likelihood:} 899.6086}\\
\multicolumn{6}{l}{$n_{Sentences}$: 2147 \qquad $n_{Speaker}$: 278} \\                   
\bottomrule
\end{tabular}
\end{subtable} \\
\begin{subtable}[c]{1\textwidth}
\centering
\subcaption{\textbf{Model 10:} \(\texttt{word\_error\_rate} \sim \texttt{country + pitch + intensity}\)}
\begin{tabular}{lrrrrr}
\toprule
\textbf{Term} & \textbf{Coef.} & \textbf{Std.Err.} & \textbf{z} & \textbf{P>\textbar z\textbar} & \textbf{[0.025, 0.975]} \\
\midrule
Intercept                & 0.475 & 0.022 & 21.826 & 0.000 & [\,0.432, 0.517\,] \\
Chile       & -0.021 & 0.017 & -1.270 & 0.204 & [\,-0.054, 0.012\,] \\
Colombia     & -0.010 & 0.017 & -0.618 & 0.537 & [\,-0.043, 0.022\,] \\
Peru      & -0.019 & 0.016 & -1.145 & 0.252 & [\,-0.051, 0.013\,] \\
Puerto Rico & -0.038 & 0.028 & -1.324 & 0.186 & [\,-0.093, 0.018\,] \\
Venezuela    & 0.019 & 0.018 &  1.063 & 0.288 & [\,-0.016, 0.053\,] \\
Pitch                    & -0.000 & 0.000 & -0.850 & 0.395 & [\,-0.000, 0.000\,] \\
Intensity                & -7.207 & 0.460 & -15.669 & 0.000 & [\,-8.109, -6.306\,] \\
Group Var                & 0.003 & 0.005 &        &       &                     \\
\midrule
\multicolumn{6}{l}{\textbf{AIC:} -1779.5362 \quad \textbf{BIC:} -1724.2260 \quad
                   \textbf{Log-Likelihood:} 899.7681}\\
\multicolumn{6}{l}{$n_{Sentences}$: 2147 \qquad $n_{Speaker}$: 278} \\
\bottomrule
\end{tabular}
\end{subtable}
\end{table}

\begin{table}
\ContinuedFloat 
\centering
\begin{subtable}[c]{1\textwidth}
\centering
\subcaption{\textbf{Model 11:} \(\texttt{word\_error\_rate} \sim \texttt{gender + pitch + intensity + pitch:gender}\)}
\begin{tabular}{lrrrrr}
\toprule
Intercept             & 0.456 & 0.030 & 15.040 & 0.000 & [\,0.397, 0.516\,] \\
Male        & 0.002 & 0.055 &  0.029 & 0.977 & [\,-0.106, 0.109\,] \\
Pitch                 & -0.000 & 0.000 & -0.389 & 0.697 & [\,-0.000, 0.000\,] \\
Pitch:Male  & 0.000  & 0.000 &  0.074 & 0.941 & [\,-0.001, 0.001\,] \\
Intensity             & -7.100 & 0.458 & -15.488 & 0.000 & [\,-7.999, -6.202\,] \\
Group Var             & 0.002 & 0.004 &        &       &                     \\
\midrule
\multicolumn{6}{l}{\textbf{AIC:} -1779.0667 \quad \textbf{BIC:} -1740.3496 \quad
                   \textbf{Log-Likelihood:} 896.5334}\\
\multicolumn{6}{l}{$n_{Sentences}$: 2147 \qquad $n_{Speaker}$: 278} \\
\bottomrule
\end{tabular} 
\end{subtable} \\
\begin{subtable}[c]{1\textwidth}
\centering
\subcaption{\textbf{Model 12:} \(\texttt{word\_error\_rate} \sim \texttt{country + gender + pitch + intensity + pitch:gender}\)}
\begin{tabular}{lrrrrr}
\toprule
\textbf{Term} & \textbf{Coef.} & \textbf{Std.Err.} & \textbf{z} & \textbf{P>\textbar z\textbar} & \textbf{[0.025, 0.975]} \\
\midrule
Intercept                & 0.473 & 0.032 & 14.924 & 0.000 & [\,0.411, 0.535\,] \\
Chile       & -0.022 & 0.017 & -1.285 & 0.199 & [\,-0.055, 0.011\,] \\
Colombia     & -0.010 & 0.017 & -0.614 & 0.539 & [\,-0.043, 0.022\,] \\
Peru      & -0.019 & 0.016 & -1.168 & 0.243 & [\,-0.051, 0.013\,] \\
Puerto Rico & -0.037 & 0.029 & -1.289 & 0.198 & [\,-0.093, 0.019\,] \\
Venezuela    & 0.018 & 0.018 &  1.041 & 0.298 & [\,-0.016, 0.053\,] \\
Male           & -0.009 & 0.056 & -0.159 & 0.874 & [\,-0.119, 0.101\,] \\
Pitch                    & -0.000 & 0.000 & -0.525 & 0.599 & [\,-0.000, 0.000\,] \\
Pitch:Male     & 0.000  & 0.000 &  0.224 & 0.823 & [\,-0.001, 0.001\,] \\
Intensity                & -7.218 & 0.471 & -15.330 & 0.000 & [\,-8.141, -6.295\,] \\
Group Var                & 0.003 & 0.005 &        &       &                     \\
\midrule
\multicolumn{6}{l}{\textbf{AIC:} -1776.0150 \quad \textbf{BIC:} -1709.6428 \quad
                   \textbf{Log-Likelihood:} 900.0075}\\
\multicolumn{6}{l}{$n_{Sentences}$: 2147 \qquad $n_{Speaker}$: 278} \\
\bottomrule
\end{tabular}
\end{subtable}
\end{table}